\documentclass[10pt,twocolumn,letterpaper]{article}

\usepackage{iccv}
\usepackage{times}
\usepackage{graphicx}
\usepackage{amsmath}
\usepackage{amssymb}
\usepackage{authblk}
\usepackage{epstopdf}
\usepackage{multirow}

\iccvfinalcopy 

\def\iccvPaperID{****} 

\ificcvfinal\fi

\begin{document}

\title{DeepMark: One-Shot Clothing Detection}
\author[ ]{
	Alexey Sidnev\textsuperscript{1,2},
	Alexey Trushkov\textsuperscript{1},
	Maxim Kazakov\textsuperscript{1,3},
	Ivan Korolev\textsuperscript{1},
	Vladislav Sorokin\textsuperscript{1}
}
\affil[1]{Huawei Research Center, Nizhny Novgorod, Russia}
\affil[2]{Lobachevsky State University of Nizhny Novgorod, Russia}
\affil[3]{National Research University Higher School of Economics, Nizhny Novgorod, Russia}
\affil[ ]{\textit {\{sidnev.alexey, trushkov.alexey, kazakov.maxim, korolev.ivan, sorokin.vladislav\}@huawei.com}}
\maketitle
\ificcvfinal\thispagestyle{empty}\fi

\begin{abstract}
The one-shot approach, DeepMark, for fast clothing detection as a modification of a multi-target network, CenterNet \cite{CenterNet}, is proposed in the paper. The state-of-the-art accuracy of 0.723 $\mathit{mAP}$ for bounding box detection task and 0.532 $\mathit{mAP}$ for landmark detection task on the DeepFashion2 Challenge dataset \cite{DF2Challenge} were achieved. The proposed architecture can be used effectively on the low-power devices.
\end{abstract}

\section{Introduction}
Fashion image analysis and processing are highly demanded tasks and there are some popular services that require it \cite{Syte}, \cite{Vue}, \cite{Oyper}, \cite{Looklet}. For example, virtual fitting rooms are becoming very common \cite{WannaKicks}, \cite{Glamstorm}.

Fashion analysis is quite a complex task. It requires solving of the following problems: detection, fine-gained classification (attributes), image retrieval, and overlay (augmented reality). Clothing detection is the first technological challenge to be solved for almost any fashion analysis task. Identifying the location of clothing items can be done with the different levels of precision: bounding boxes, keypoints, or segmentation. The first two tasks can provide enough information for the most of the real-life use-cases.

Data safety policies and customer needs arise another challenge requiring all processing to be done on the low-power edge devices like mobile phones and smart cameras.
In accordance with the proposed approach, the tasks of bounding box prediction and keypoint estimation can be performed with a single one-shot neural network.
It can also be used effectively on low-power devices without any post-processing work needed.

\section{Related work}

\begin{figure}[ht]
	\centering
	\includegraphics[width=\columnwidth]{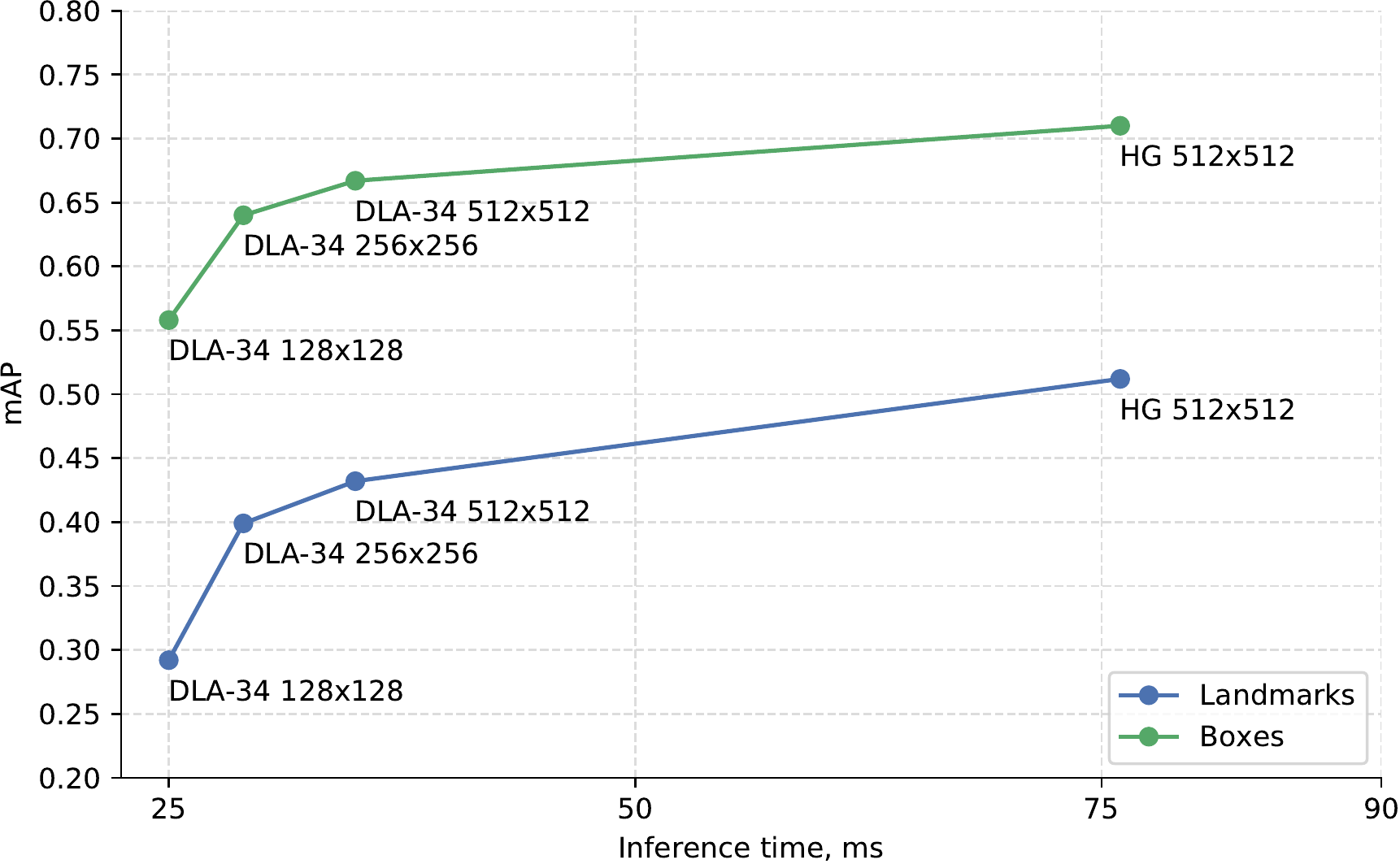}
	\caption{Speed-accuracy trade-off for object detection and landmark estimation on DeepFashion2 Challenge dataset \cite{DF2Challenge}. We considered different image resolutions: $128 \times 128$, $256 \times 256$ and $512 \times 512$. DLA-34 model was initially trained on $512 \times 512$ images and then fine-tuned for other resolutions. Inference time was evaluated without flip and multiscale post processing.}
	\label{fig:plot}
\end{figure}

Visual analysis of fashion images includes a variety of tasks: fashion attribute prediction; clothing recognition, detection and segmentation; estimation of fashion landmarks, clothing image retrieval and identification of fashion trends~\cite{wang2011clothes}, \cite{chen2012describing}, \cite{zheng2018modanet}, \cite{kiapour2014hipster}, \cite{liu2016deepfashion}. Previously, the different ways of distinctive features capturing in the fashion domain have been researched: full image usage \cite{hadi2015buy}, general object proposal \cite{hadi2015buy}, bounding boxes \cite{chen2015deep}, masks \cite{liang2015human}, fashion attributes \cite{liu2016deepfashion} and fashion landmarks \cite{liu2016fashion}.

Recent research shows that fashion landmarks are the most efficient, discriminative and robust representation for visual fashion analysis \cite{liu2016deepfashion}, \cite{liu2016fashion}.

Recent CNN-based approaches \cite{liu2016deepfashion}, \cite{liu2016fashion}, \cite{lu2017fully}, \cite{liang2015human}, \cite{huang2015cross} along with the advent of large-scale fashion datasets \cite{liu2016deepfashion}, \cite{zheng2018modanet} have significantly outperformed the prior work. The success of those deep learning based fashion models has demonstrated the strong representative power of neural networks and the advantages of fashion landmarks as features for fashion analysis \cite{liu2016fashion}.

However, until the release of the newest DeepFashion2~\cite{DeepFashion2} dataset the full richness of domain-specific clothing knowledge had remained unresearched. Previously, the largest existing fashion dataset DeepFashion \cite{liu2016deepfashion} was limited by a single clothing item per image, few landmarks and pose definitions.

DeepFashion2 \cite{DeepFashion2} addressed such drawbacks providing a large-scale benchmark with comprehensive tasks and annotations. DeepFashion2 enabled the development of more advanced algorithms for visual fashion analysis.

In the paper, the tasks of clothing landmark prediction and clothing detection using DeepFashion2 Challenge \cite{DF2Challenge} dataset are focused on.

The previous methods in the area in question describe detection of joints and other landmarks for facial alignment \cite{ramanan2012face} and estimation of human body poses \cite{tompson2014joint}. There are also some new works which followed the keypoint estimation approach for object detection. CornerNet \cite{law2018cornernet} detects two bounding box corners as keypoints while ExtremeNet~\cite{zhou2019bottom} detects the top-, left-, bottom-, rightmost; and center points of all objects. There is also CenterNet, which extracts a single center point per object \cite{CenterNet}.

\section{Proposed approach}

Due to the lack of existing works in fashion landmark detection, approaches from the similar fields can be good research-subjects (e.g. multi-person pose estimation \cite{CenterNet}, \cite{cao2018openpose}, \cite{MaskRCNN}, \cite{AssociativeEmbedding}, \cite{PersonLab}). Recently, CenterNet \cite{CenterNet} has achieved the best speed-accuracy trade-off in object detection and pose estimation tasks on the MS COCO dataset. It verifies that CenterNet is a general approach that can be adapted to a new task, namely to fashion landmark detection. 
The primary challenge resides in fact, that CenterNet has been trained and has demonstrated solid results on the MS COCO pose estimation dataset, which includes only one class of objects (human) and a fixed number of keypoints (17). The DeepFashion2 \cite{DeepFashion2} dataset includes 13 categories each of which contains a different number of keypoints (294 keypoints total). Our approach is based on CenterNet topology but handles two above mentioned restrictions with ease. 

\begin{figure}[ht]
	\centering
	\includegraphics[width=\columnwidth]{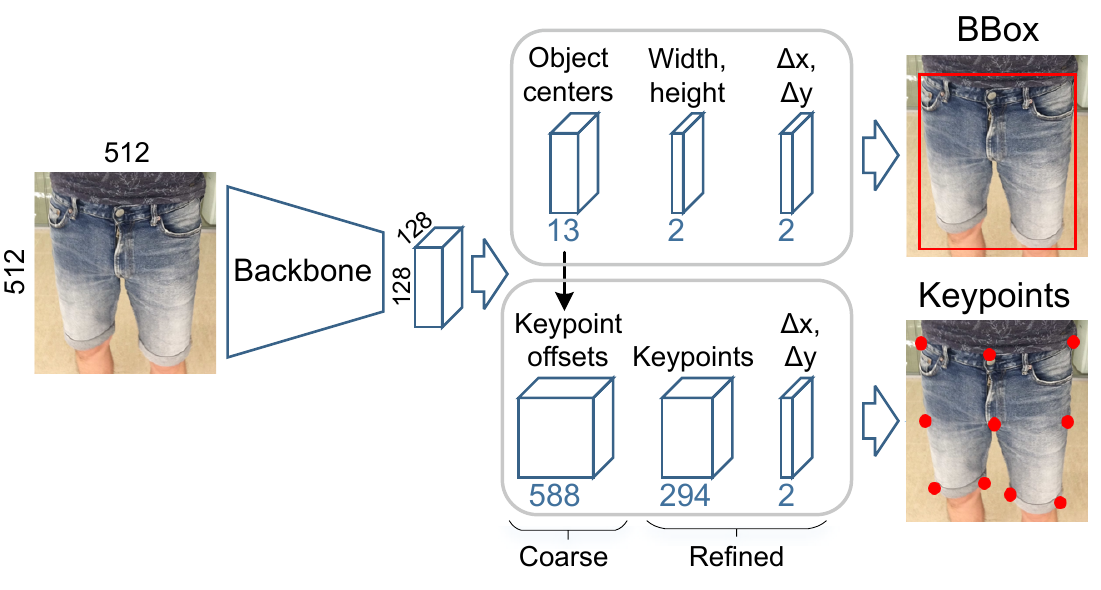}
	\caption{Scheme of DeepMark.}
	\label{fig:scheme}
\end{figure}

Fig.\ref{fig:scheme} depicts the scheme of our DeepMark approach. The DeepFashion2 dataset contains 13 classes, so 13 channels are used to predict the centers of the boxes. Width and height are predicted directly. Two more channels in the output feature map representing $\bigtriangleup$x and $\bigtriangleup$y are regressed to refine the center coordinates.

Fashion landmark estimation aims to estimate 2D keypoint locations for each item of clothing in one image. To predict the keypoints, regression-based estimator similar to~\cite{DeepPose} is used. The centers of the boxes are used to predict the coarse locations of the keypoints (keypoint offsets from Fig.\ref{fig:scheme}).
To refine the estimation of keypoints, a heatmap is generated for each of 294 keypoints. Like in the detection case, two more channels $\bigtriangleup$x and $\bigtriangleup$y are used to get more precise landmark coordinates.
In order to preserve connectivity between object center and corresponding keypoints, each refined keypoint position is assigned to its closest coarse position from the previous step. These refined coordinates represent the final results of the algorithm.

\section{Results}

All experiments were performed on the publicly available DeepFashion2 Challenge dataset \cite{DF2Challenge}, which  contains 191,961 images in the training set and 32,153 images in the validation set. The previous work \cite{DeepFashion2} used a dataset twice as big as the DeepFashion2 training set, and the validation set contained another number of images. More details are in \autoref{tab:Datasets}.

\begin{table}
	\begin{center}
		\begin{tabular}{|c|c|c|c|}
			\hline
			Subset & Challenge  & Paper \\
			\hline			
			Train & 191961 & 390884 \\
			\hline
			Val & 32153 & 33669 \\
			\hline
		\end{tabular}
	\end{center}
	\caption{Training and validation subsets provided for DeepFashion2 Challenge \cite{DF2Challenge} and used in original paper \cite{DeepFashion2}.}
	\label{tab:Datasets}	
\end{table}

\begin{table}
	\begin{center}
		\begin{tabular}{|c|c|c|c|}
			\hline
			Metric & DeepMark & Challenge & Paper \\
			\hline
			\multirow{2}{*}{$\mathit{mAP}_{pt}$} 
			& 0.601 & {\bf0.605} & 0.641 \\
			& {\bf0.532} & 0.529 & 0.563 \\
			\hline
			\multirow{2}{*}{$\mathit{mAP}^{iou=0.50}_{pt}$} 
			& {\bf0.793} & 0.790 & 0.820 \\
			& {\bf0.784} & 0.775 & 0.805 \\
			\hline
			\multirow{2}{*}{$\mathit{mAP}^{iou=0.75}_{pt}$} 
			& 0.683 & {\bf0.684} & 0.728 \\
			& {\bf0.599} & 0.596 & 0.641 \\
			\hline
		\end{tabular}
	\end{center}
	\caption{Landmark estimation results for DeepMark and Mask R-CNN \cite{DeepFashion2}. The proposed approach outperforms DeepFashion2 Challenge \cite{DF2Challenge} baseline on the almost all metrics. Results of evaluation on visible landmarks only and evaluation on both visible and occluded landmarks are separately shown in each row.}
	\label{tab:LandmarkCompare}	
\end{table}

\begin{table}
	\begin{center}
		\begin{tabular}{|c|c|c|}
			\hline
			Metric & DeepMark & Paper \\
			\hline
			$\mathit{mAP}_{box}$ & {\bf0.723} & 0.667 \\
			\hline			
			$\mathit{mAP}^{iou=0.50}_{box}$ & {\bf0.827} & 0.814 \\
			\hline			
			$\mathit{mAP}^{iou=0.75}_{box}$ & {\bf0.795} & 0.773 \\
			\hline
		\end{tabular}
	\end{center}
	\caption{Clothing detection results for DeepMark and Mask R-CNN \cite{DeepFashion2}. The proposed model significantly outperforms Mask R-CNN while been trained on half of the images.}
	\label{tab:DetectionCompare}		
\end{table}

DeepMark has achieved the best results using the 8 GPU system with Nvidia GeForce RTX 2080ti graphics card after training the model for 63 epochs on the DeepFashion2 Challenge training set (191,961 images) with a batch of 37 images and input network resolution of $512 \times 512$. We used the CenterNet MS COCO model for object detection as the initial checkpoint and have performed experiments with two backbones: Hourglass and DLA-34. Learning rate schedule: 3e-4 - 50 epochs, 3e-5 - 3 epochs, 1e-5 - 10 epochs.

DeepMark outperformed the DeepFashion2 Challenge~\cite{DF2Challenge} baseline on the almost all metrics (\autoref{tab:LandmarkCompare}). DeepFashion2 Challenge only provides a baseline for the landmark estimation problem, so 
comparison only with the paper \cite{DeepFashion2} was provided for this task in \autoref{tab:DetectionCompare}.

\begin{table}
    \begin{center}
        \begin{tabular}{ |c|c|c|c|c| } 
            \hline
            NMS & - & \checkmark & \checkmark & \checkmark \\ 
            Flip & - & - & \checkmark & \checkmark \\ 
            Multiscale & - & - &- & \checkmark \\
            \hline
            $\mathit{mAP}_{box}$ & 0.707 & 0.710 & 0.723 & {\bf0.723}\\ 
            $\mathit{mAP}_{pt}$ & 0.511 & 0.512 & 0.529 & {\bf0.532}\\ 
            \hline
            Inference time, ms & 73 & 76 & 165 & 315\\ 
            \hline
        \end{tabular}
    \end{center}
	\caption{Comparison of different post-processing strategies with Hourglass backbone.}
	\label{tab:Postprocessing}
\end{table}

\begin{table}
	\begin{center}
		\begin{tabular}{|c|c|c|c|c|}
			\hline
			\multirow{2}{*}{ }                        
			& \multicolumn{2}{ |c| }{DLA-34} & \multicolumn{2}{ |c| }{Hourglass} \\\cline{2-5}
			& Single & Fusion & Single & Fusion \\
			\hline
			$\mathit{mAP}_{box}$ & 0.667 & 0.686 & 0.710 & 0.723 \\
			\hline
			$\mathit{mAP}^{iou=0.50}_{box}$ & 0.788 & 0.802 & 0.821 & 0.827 \\
			\hline
			$\mathit{mAP}^{iou=0.75}_{box}$ & 0.750 & 0.767 & 0.786 & 0.795 \\
			\hline
			Inference time, ms & 35 & 216 & 76 & 315 \\
			\hline
		\end{tabular}
	\end{center}
	\caption{Clothing detection.}
	\label{tab:DetectionResults}		
\end{table}

\begin{table}
    \begin{center}
        \begin{tabular}{|c|c|c|c|c|} 
            \hline
            \multirow{2}{*}{ }                        
             & \multicolumn{2}{ |c| }{DLA-34} & \multicolumn{2}{ |c| }{Hourglass} \\\cline{2-5}
             & Single & Fusion & Single & Fusion \\
            \hline
            \multirow{2}{*}{$\mathit{mAP}_{pt}$} 
             & 0.469 & 0.513 & 0.582 & 0.601 \\
             & 0.432 & 0.448 & 0.512 & 0.532 \\
            \hline
            \multirow{2}{*}{$\mathit{mAP}^{oks=0.50}_{pt}$} 
            & 0.730 & 0.746 & 0.784 & 0.793 \\
            & 0.716 & 0.730 & 0.774 & 0.784 \\
            \hline
            \multirow{2}{*}{$\mathit{mAP}^{oks=0.75}_{pt}$} 
            & 0.547 & 0.563 & 0.660 & 0.683 \\
            & 0.455 & 0.473 & 0.571 & 0.599 \\
            \hline
            Inference time, ms & 35 & 216 & 76 & 315 \\
            \hline
        \end{tabular}
    \end{center}
	\caption{Clothing landmark estimation.}
	\label{tab:LandmarkResults}	
\end{table}

Different inference strategies are considired in the paper including applying non-maximum suppression, fusing model outputs from original and flipped images with equal weighs and from original images; and images downscaled by 0.75. The results of these strategies for landmark estimation and object detection problems are shown in \autoref{tab:Postprocessing}. The mixture of all inference strategies adds about 2 mAP over the vanilla solution.

\autoref{tab:DetectionResults} and \autoref{tab:LandmarkResults} include results for object detection and landmark estimation problems for visible and visible+occluded landmarks. The DLA-34 model provides the best trade-off between speed and accuracy which makes it a good choice for the low-power devices. It works in only 35 ms per image on a single GPU, with an accuracy of 0.432 $\mathit{mAP}_{pt}$ and 0.667 $\mathit{mAP}_{box}$. With the Hourglass model we have achieved the state-of-the-art results for both tasks.

\section{Conclusion}
The new approach, DeepMark, is proposed as an adaptation of CenterNet \cite{CenterNet} for clothing landmark estimation tasks. The state-of-the-art accuracy was achieved on the DeepFashion2 dataset for 2 tasks: clothing detection (0.723 $\mathit{mAP}$) and clothing landmark estimation (0.532 $\mathit{mAP}$). The proposed neural network architecture has just one stage. It can be used without post-processing. It takes 35 ms per image and has quite high accuracy (0.432 $\mathit{mAP}_{pt}$ and 0.667 $\mathit{mAP}_{box}$) for clothing detection tasks (Fig.\ref{fig:plot}).

{\small
\bibliographystyle{ieee}
\bibliography{egbib}
}

\begin{figure}[t]
	\centering
	\includegraphics[width=\columnwidth]{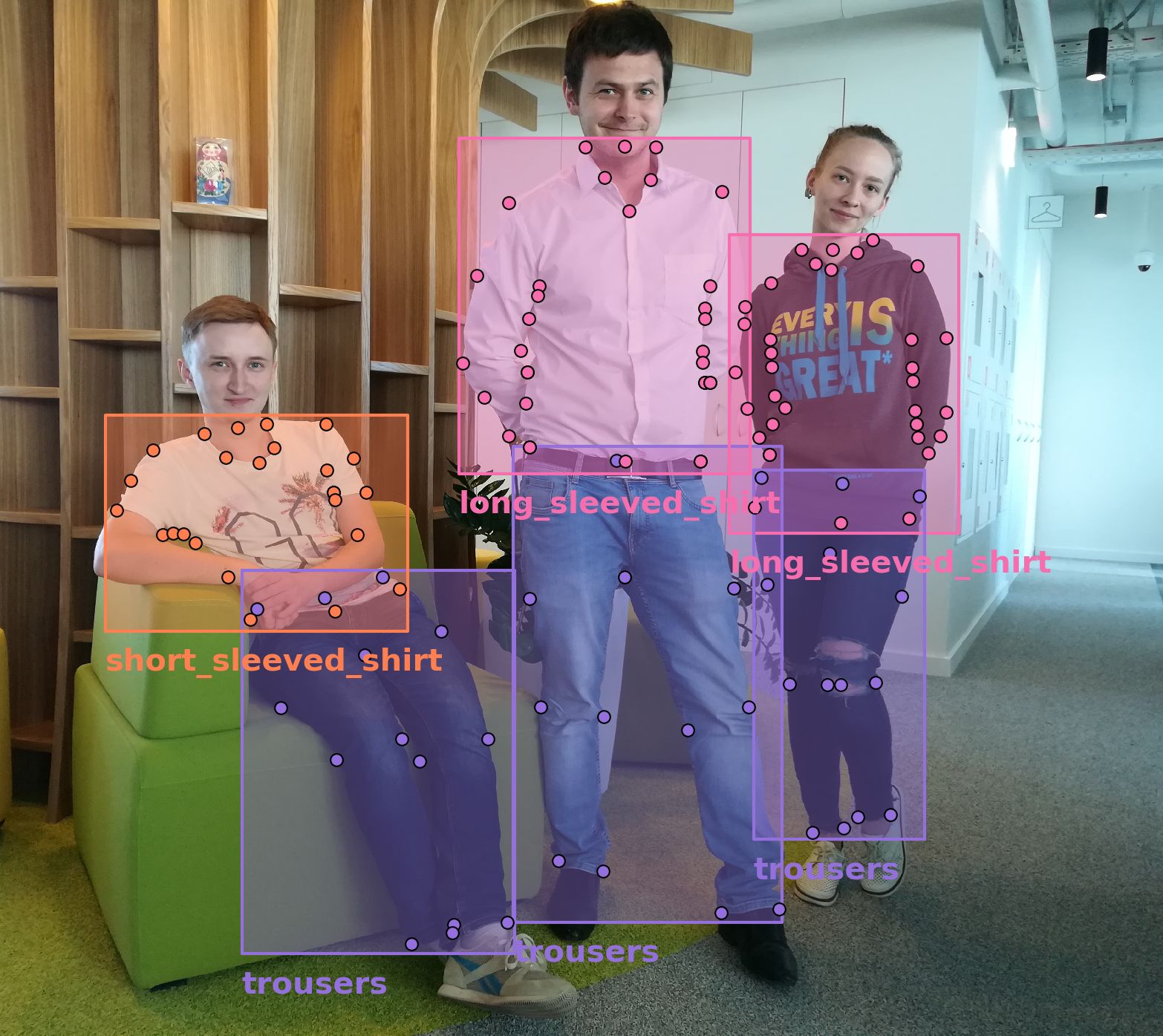}
	\caption{Example of DeepMark results. Best viewed in color.}
	\label{fig:team}
\end{figure}

\end{document}